\documentclass{article}
\usepackage{amssymb}
\usepackage{wrapfig}
\usepackage{array}
\usepackage{adjustbox}
\usepackage[preprint]{corl_2025}
\usepackage{subcaption}
\usepackage[numbers]{natbib}
\usepackage{amsmath}
\usepackage{booktabs}
\usepackage{float}
\usepackage{graphicx}
\usepackage{multirow}
\usepackage{url}
\usepackage[table]{xcolor}
\usepackage{enumitem} 

\usepackage[font=footnotesize, labelformat=simple]{subcaption}

\newcommand{\bench}{\textit{Any}Body}
\newcommand{\reach}{\texttt{reach}}
\newcommand{\push}{\texttt{push}}

\setlist[itemize]{ 
  leftmargin=2.5em,
  labelsep=0.5em,  
  itemsep=0.2em,   
}

\setlist[enumerate]{ 
  leftmargin=2.5em,  
  labelsep=0.5em,    
  itemsep=0.2em,     
}

\author{
    Meenal Parakh, \;
    Alexandre Kirchmeyer, \;
    Beining Han, \;
    Jia Deng \\
\texttt{\{meenalp,  akirchmeyer, bh7032, jiadeng\}@princeton.edu} \\ 
Princeton University
}

\title{\bench: A Benchmark Suite for Cross-Embodiment Manipulation }

\begin{document}
\maketitle


\begin{abstract}
Generalizing control policies to novel embodiments remains a fundamental challenge in enabling scalable and transferable learning in robotics. While prior works have explored this in locomotion, a systematic study in the context of manipulation tasks remains limited, partly due to the lack of standardized benchmarks. In this paper, we introduce a benchmark for learning cross-embodiment manipulation, focusing on two foundational tasks—reach and push—across a diverse range of morphologies. The benchmark is designed to test generalization along three axes: interpolation (testing performance within a robot category that shares the same link structure), extrapolation (testing on a robot with a different link structure), and composition (testing on combinations of link structures). On the benchmark, we evaluate the ability of different RL policies to learn from multiple morphologies and to generalize to novel ones. Our study aims to answer whether morphology-aware training can outperform single-embodiment baselines, whether zero-shot generalization to unseen morphologies is feasible, and how consistently these patterns hold across different generalization regimes. The results highlight the current limitations of multi-embodiment learning and provide insights into how architectural and training design choices influence policy generalization. Project page: 
\href{https://princeton-vl.github.io/anybody/}{\texttt{https://princeton-vl.github.io/anybody}.}

\end{abstract}

\section{Introduction}

Generalizing control policies across diverse embodiments is a fundamental challenge in robotics, with broader implications for building agents that exhibit \textit{general intelligence}. Humans naturally exhibit this ability—we can easily infer how to operate tools and machines of various shapes and functionalities, from robotic arms and mobile manipulators to coffee machines, cars, and arcade claw machines. This capacity to reason over varied action spaces and control diverse embodiments is a key aspect of general intelligence.

Alongside its role in general intelligence, cross-embodiment learning has practical benefits in terms of \textit{scalability} and \textit{transferability}. It allows scaling up robot training by leveraging large-scale, heterogeneous datasets (e.g., Open-X \cite{ONeill2024OpenXR}, DROID \cite{Khazatsky2024DROIDAL}) to enable deployment across different lab environments or robotic platforms without the need for extensive retraining.

Despite these advantages, a central factor in cross-embodiment learning—robot morphology—is often overlooked. In many large-scale training setups, explicit morphology information is excluded from the inputs, leaving open questions about the potential benefits. In particular, incorporating morphology could enable zero-shot generalization to unseen embodiments \cite{Gupta2022MetaMorphLU, patel2024getzero, Yang2024PushingTL}, offering an alternative to the common reliance on fine-tuning \cite{Kim2024OpenVLAAO, Team2024OctoAO}. While several works explore morphology-aware approaches \cite{Chen2018HardwareCP, patel2024getzero, Hu2021KnowTT}, their evaluations are often tailored to the specific methods, or focus specifically on locomotion \cite{Gupta2022MetaMorphLU, sferrazza2024body, kurin2021my, Wang2018NerveNetLS}. The absence of standardized evaluations makes it difficult to measure progress and compare cross-embodiment learning methods in manipulation.

Evaluation setups in existing methods often involve only a limited number of robots--7 in \cite{Chen2018HardwareCP}, and 4 in \cite{Hu2021KnowTT, Yang2023PolybotTO}-- and typically focus on morphologies with similar affordances, such as variations of robotic arms or multi-fingered hands \cite{patel2024getzero}.
To enable more rigorous evaluation of cross-embodiment learning, a benchmark should (a) include a broad range of robot morphologies with clearly defined train and test splits, and (b) reflect the core challenge of reasoning about morphology and the affordances it enables.
\begin{figure}
    \centering
    \includegraphics[width=\linewidth]{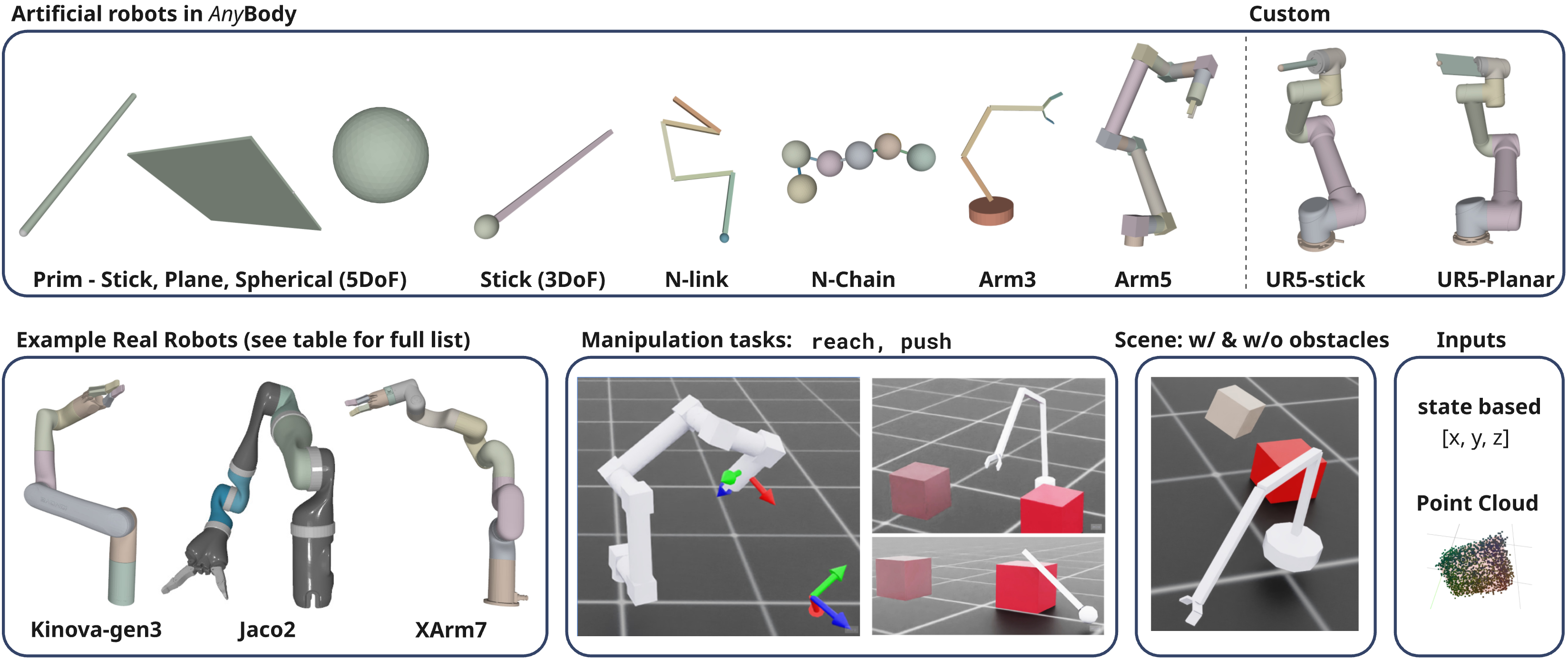}
    \caption{\footnotesize We introduce \bench, a benchmark suite for evaluating policy generalization across diverse robot morphologies. It consists of 18 robot variations: 8 procedurally generated robot categories and 10 based on real-world robots. The benchmark tasks comprise two manipulation tasks —\reach~and \push — two scene variations (with and without obstacles), and two input types—state-based and point cloud-based. }
    \label{fig:task}
    \vspace*{-10pt}
\end{figure}

To address this need, we introduce \bench, a suite of simulated environments designed to evaluate policy generalization across diverse robot morphologies. The benchmark focuses on two core manipulation tasks— \reach~and \push—which have been widely used in prior work \cite{Chen2018HardwareCP, Hu2021KnowTT, yu2021metaworldbenchmarkevaluationmultitask}. We find that these tasks are challenging and well-suited for assessing the generalization we aim to study.

\bench~includes both simple and complex robots that capture a wide range of morphological diversity and affordances (Figure \ref{fig:task}). Simple robots test a model’s ability to learn basic capabilities, while complex robots assess its ability to handle richer, multi-joint control like reaching. This range ensures methods can generalize across affordances and scale with morphological complexity, while avoiding bias toward high-DOF arms by including minimal, abstract morphologies that challenge agents to reason from first principles.

\begin{wrapfigure}{r}{0.42\textwidth}
  \centering
  \vspace*{-20pt}
    \includegraphics[width=0.42\textwidth]{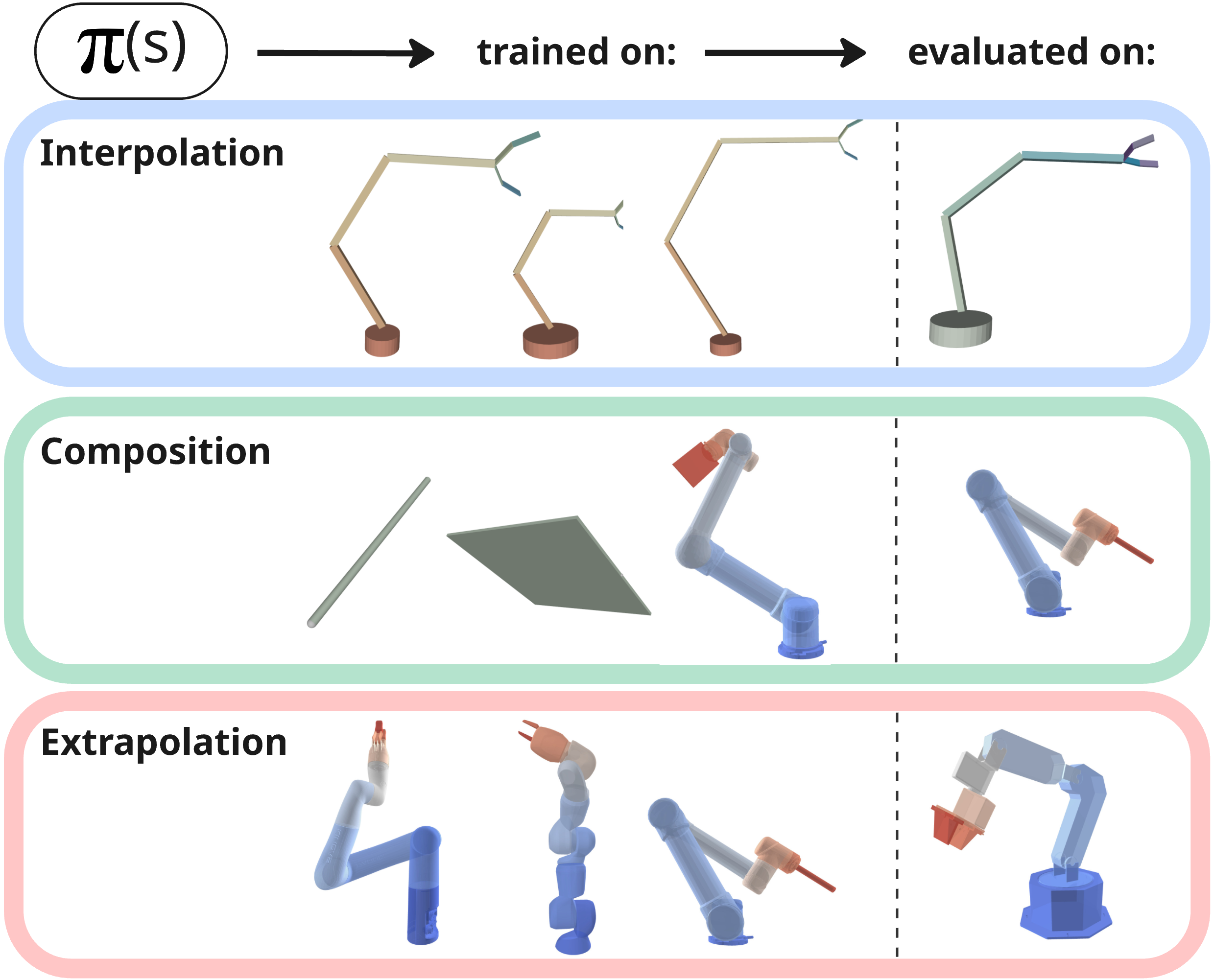}
  \vspace*{-15pt}
  \caption{\footnotesize Three categories of benchmark tasks. We aim to test the zero-shot generalization ability of a multi-embodiment policy $\pi(s)$ on unseen morphologies.}
  \vspace*{-10pt}
  \label{fig:bench_tasks}
\end{wrapfigure}

Framed as a multi-task learning problem, the benchmark evaluates a method’s ability to learn from multi-embodiment data and generalize to novel morphologies. It tests generalization along three axes: interpolation, extrapolation, and composition (see Figure \ref{fig:bench_tasks}).
In the \textit{interpolation} setting, agents are trained and tested on robots within the same category, but with geometrical variations. In the \textit{extrapolation} setting, agents are trained on multiple robot categories and evaluated on a robot with a different link structure. Finally, in the \textit{composition} setting, the test morphology is composed of components seen during training but assembled in a new configuration. This setting tests whether agents can reason about the functionality of individual parts and combine them to infer the behavior of the whole robot—a key aspect of general intelligence. The composition setting, in particular, is underexplored and introduces a new dimension for evaluating policy generalization.

On these benchmark tasks, our evaluations with RL agents aim to investigate several critical aspects: whether morphology-aware training can outperform single-embodiment baselines, the feasibility of zero-shot generalization to unseen morphologies, and how these patterns hold across different generalization regimes. The study highlights key challenges in multi-embodiment learning and offers insights into how architectural and training design choices affect policy generalization.

Our key results confirm that the benchmark includes tasks of varying difficulty, enabling fine-grained analysis of cross-embodiment generalization along the morphology axis. Further, we show that while in-distribution generalization is feasible, zero-shot generalization in extrapolation and composition settings remains challenging.

In summary, our key contributions are:
\begin{enumerate}
\item We introduce \bench, a benchmark for evaluating the generalizability of robotic manipulation policies to novel embodiments. With the benchmark, we provide an open-source codebase that extends IsaacSim \cite{isaacsim} with multi-task training capabilities—a feature not currently supported by the IsaacLab wrapper \cite{isaaclab, isaaclab2022}.
\item We present a systematic evaluation of RL agents on this benchmark, identifying key challenges and effective design choices for multi-embodiment learning.
\end{enumerate}

\section{Related Works}

\textbf{Cross-Embodiment Datasets.} Large-scale cross-embodiment datasets, such as \cite{ONeill2024OpenXR, Khazatsky2024DROIDAL}, contain data collected across various embodiments. Several works \cite{ONeill2024OpenXR, bjorck2025gr00t, black2024pi0, kim2024openvla, team2024octo} have trained large-scale vision-language-action models on these cross-embodiment datasets, achieving impressive few-shot and even zero-shot performance on unseen embodiments. However, since the evaluation robots are typically included in the pretraining datasets, it becomes difficult to assess the models' true generalization capabilities. In addition, methods such as \cite{Yang2024PushingTL, bousmalis2023robocat, Reed2022AGA}, trained on a diverse range of robots, often introduce perception challenges. Our benchmark aims to address these limitations and systematically test agents' ability to generalize to different morphologies.

\textbf{Simulation Benchmarks.} The robotics community has developed several simulation benchmarks, ranging from simple manipulation skills \cite{gu2023maniskill2, ehsani2021manipulathor, james2020rlbench, zhu2020robosuite, yu2020meta, heo2023furniturebench, han2024fetchbench} to complex, long-horizon tasks \cite{li2023behavior1k, srivastava2022behavior100, szot2021habitat2}, and even humanoid robots \cite{sferrazza2024humanoidbench}. However, these benchmarks typically focus on a limited set of robots with minimal variation in morphology \cite{zhu2020robosuite}, and often aim to study general skill learning. In contrast, our benchmark focuses on learning skills across a range of robot morphologies. \cite{unimal} has been used in morphology-aware learning, leading to works such as \cite{Gupta2022MetaMorphLU, sferrazza2024body}; however, it is specifically for evaluating locomotion ability of agents, while we focus on learning manipulation skills. 

\textbf{Morphology-aware Learning.} Several works, such as \cite{Gupta2022MetaMorphLU, kurin2021my, Wang2018NerveNetLS, Hong2022StructureAwareTP, zhang2020crossloco}, study morphology-aware learning and typically use a multitask learning framework to train universal policies. We adopt a similar approach but focus on manipulation tasks. Research works \cite{patel2024getzero, Chen2018HardwareCP, Hu2021KnowTT, Yang2023PolybotTO} also use morphology information to learn manipulation from multi-embodiment data. However, their evaluations involve only a limited set of robot variations. In contrast, our benchmark consists of diverse robots and systematically tests cross-embodiment generalization across three axes of morphology variations.

\section{Benchmark}

\bench~aims to evaluate methods on their ability to perform manipulation tasks across a wide range of robot morphologies, and to test their generalization ability to unseen morphologies. We focus on two fundamental tasks—\reach~and \push—which capture both spatial understanding and physical interaction, and are common in prior works \cite{Chen2018HardwareCP, Hu2021KnowTT, yu2021metaworldbenchmarkevaluationmultitask}. Broadly, \reach~involves controlling a robot's joints to move its end-effector to a randomly designated target position. The \push~task involves controlling the robot's joints to move a block from left to right.

We next describe the different benchmark tasks and the problem formulation for evaluating multi-embodiment learning.

\subsection{Design}

Our benchmark design aims to capture:
\begin{enumerate}[leftmargin=0.2in, itemsep=-1pt, topsep=-1pt, partopsep=-1pt]
    \item A diverse set of robot morphologies, capturing a range of structural and functional variations.
    \item Allow testing generalization on three regimes: \textit{interpolation}, \textit{composition}, and \textit{extrapolation}-- each of which provides a different difficulty level to benchmark models.
\end{enumerate}

Figure \ref{fig:task} illustrates the different robot categories and task configurations. Table \ref{tab:bench_tasks} summarizes the different benchmark tasks in \bench. Below, we describe each of these tasks in more detail.

\begin{table}[h]
\centering
    \vspace{-0.6em}
    \caption{\footnotesize Benchmark tasks across three categories.}
    \label{tab:bench_tasks}
\resizebox{\columnwidth}{!}{
        \begin{tabular}{l|l|p{6cm}|p{3cm}}
        \toprule
        \multicolumn{2}{c|}{Benchmark Tasks} & Train Environments & Test Environment \\ 
        \midrule
        \rowcolor{cyan!2} Interpolation & Arm-3 & Arm-3 (10 parametric variations) & Arm3 (unseen) \\ 
        \rowcolor{cyan!2} & Panda & Panda (10 parametric variations) & Panda \\
         \midrule
         \rowcolor{green!2} \multirow{2}{*}{Composition} & EE Arm & Prims-plane, Prims-Cylinder, UR5-Planar, UR5-Ez, UR5-Sawyer & UR5-Stick \\ 
        \rowcolor{green!2}  & EE-Task & Prims-Plane (\push), UR5-Stick (\reach) & UR5-Planar (\push) \\ 
        \midrule
        \rowcolor{red!2} Extrapolation& Primitives & Stick, NLink, Prims & Chain \\ 
        \rowcolor{red!2} & Robot Arms & UR5-stick, UR5-Ez, Panda, Kinova, Jaco, XArm, LWR, Yumi, Arm5 & WidowX \\ 
        \bottomrule
        \end{tabular}
        }
\end{table}

\textbf{1. Interpolation: Generalization Within a Morphology Category.} This setting tests whether agents can generalize to new morphologies that share the same morphological structure as the training robots but differ in geometry. We evaluate both \reach~and \push~in this category.
\begin{itemize}[leftmargin=0.3in, itemsep=-1pt, topsep=-1pt, partopsep=-1pt]
\item \textbf{Arm3}: A 3-DOF arm with varying link lengths and widths generated procedurally. The test is a held-out morphological variation of \textit{Arm3}.
\item \textbf{Panda}: Modified versions of the Franka Panda arm with scaled link dimensions. We use the original Franka arm for testing. 
\end{itemize}

\textbf{2. Composition: Generalization via Recombination of Known Components.} This setting tests whether agents can reason about individual robot components and generalize to new combinations, signifying a compositional understanding of morphology.
\begin{itemize}[leftmargin=0.3in, itemsep=-1pt, topsep=-1pt, partopsep=-1pt]
    \item \textbf{EE-Arms}: Train robots consist of standard arms and various end-effectors (e.g., gripper, stick, plane). The test robot combines an arm with a novel end-effector from the train set, but unseen as a whole.
    \item \textbf{EE-Task Transfer}: A more challenging multi-embodiment, multi-task setup. Agents are trained on \textit{prims-plane} for \push~and \textit{ur5-stick} for \reach. At test time, the robot is \textit{ur5-plane}, requiring the agent to combine the pushing ability of the plane with the control capabilities of the UR5.
\end{itemize}

\textbf{3. Extrapolation: Generalization Across Morphology Categories.}
This setting evaluates whether agents trained on multiple morphology variations (possibly different link structures) can generalize to a novel robot structure. Both \reach~and \push~are used to evaluate this category.
\begin{itemize}[leftmargin=0.3in, itemsep=-1pt, topsep=-1pt, partopsep=-1pt]
    \item \textbf{Primitives}: Consists of simplified morphologies—such as \textit{stick}, \textit{n-link}, and \textit{prims}—used during training. The test robot, \textit{chain}, is structurally more complex, combining elements of the training robots (e.g., chain-like structure similar to \textit{n-link} robots, and joint structure of \textit{stick}, and built from spherical \textit{primitives} geometry).
    \item \textbf{Robot Arms}: Includes several arms from the real world with differing link structures and joint types. The test robot is another real-world arm with a previously unseen configuration.
\end{itemize}

\begin{figure}
  \centering
    \includegraphics[width=0.6\textwidth]{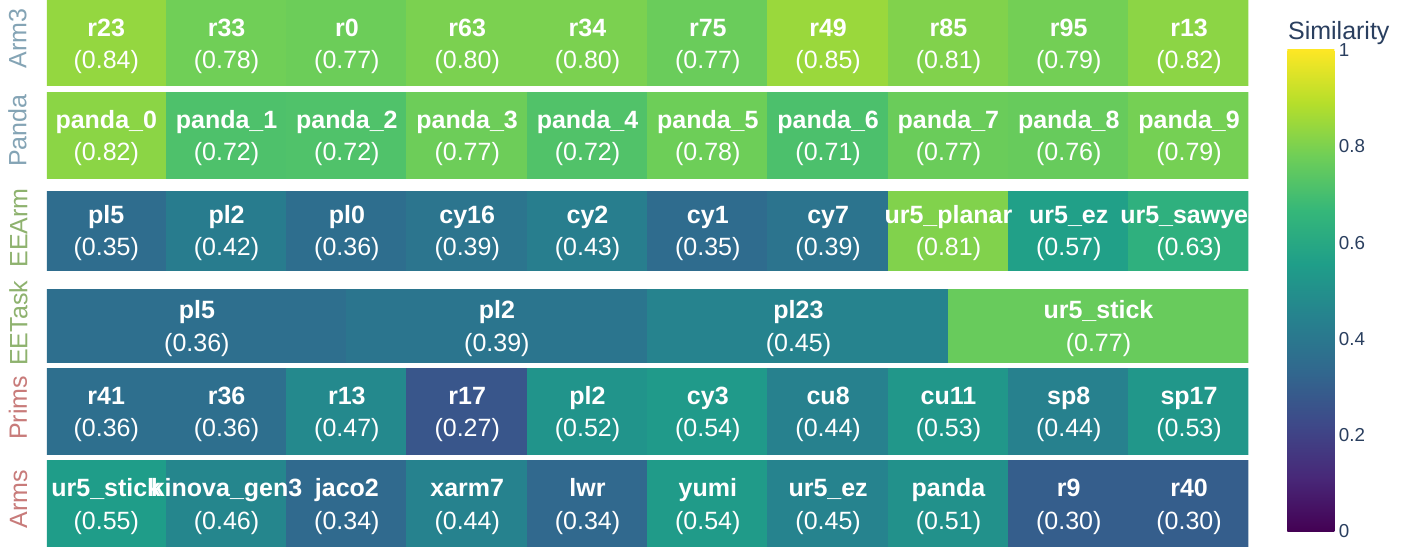}
  \caption{\footnotesize Cosine similarity of test morphologies with those in the train set.}
  \vspace*{-10pt}
  \label{fig:cosine_dist}
\end{figure}

\vspace{5pt}
\textbf{Cosine similarity.} Figure \ref{fig:cosine_dist} showcases the cosine similarity $\cos(v_{test}, v_{train})$ where $v_{test}$ and $v_{train}$ are the vector representations (discussed in next section) of the test and train morphologies, respectively. From the cosine similarities, we can observe that in the interpolation category, the test robot is closer to the train robots (and within the two interpolations, one is slightly harder than the other). For composition, some of the train robots have high similarity with the test robot, and for extrapolation, we observe that the test robot is farther from the train robots. 

\subsection{Modeling}
Following prior works \cite{Gupta2022MetaMorphLU, Chen2018HardwareCP, kurin2021my}, we model the different robotic environments as sets of infinite horizon, discounted Markov Decision Processes (MDPs) $M = \{M_1, M_2, \cdots, M_n\}$ where $M_i$ represents the MDP for the $i$-th robot. Each MDP $M_i$ is defined as $M_i = (S_i, A_i, R_i, T_i, H, \gamma)$, where $S_i$ is the state space, $A_i$ the action space, $T_i$ the transition dynamics, and $R_i$ the reward function for the $i$-th robot. The horizon $H$ and discount factor $\gamma$ are shared across all MDPs. We aim to train a policy $\pi$ that produces actions $\mathbf{a}_t = \pi(\mathbf{s}_t)$ where $\mathbf{s}_t$ is the observation at time step $t$.  

\textbf{State Space.} The full observation at time $t$, denoted as $\mathbf{s}_t$, can be expressed as:
\[
\mathbf{s}_t = \left[ (\mathbf{s}_r)_t, (\mathbf{o}_e)_t, \mathbf{g}_r \right]
\]
where:
\( (\mathbf{s}_r)_t \) represents the robot's state at time \( t \), and  \( (\mathbf{o}_e)_t \) represents the environment observation at time \( t \). We allow for both state-based and point cloud-based representations $\mathbf{o}_e$ for environment objects (excluding the robot). For \reach, a goal token $\mathbf{g}_r$ is appended to the input sequence alongside the robot and environment states. This token is not used in \push.

\begin{wrapfigure}{r}{0.3\textwidth}
  \centering
  \vspace*{-10pt}
    \includegraphics[width=0.2\textwidth]{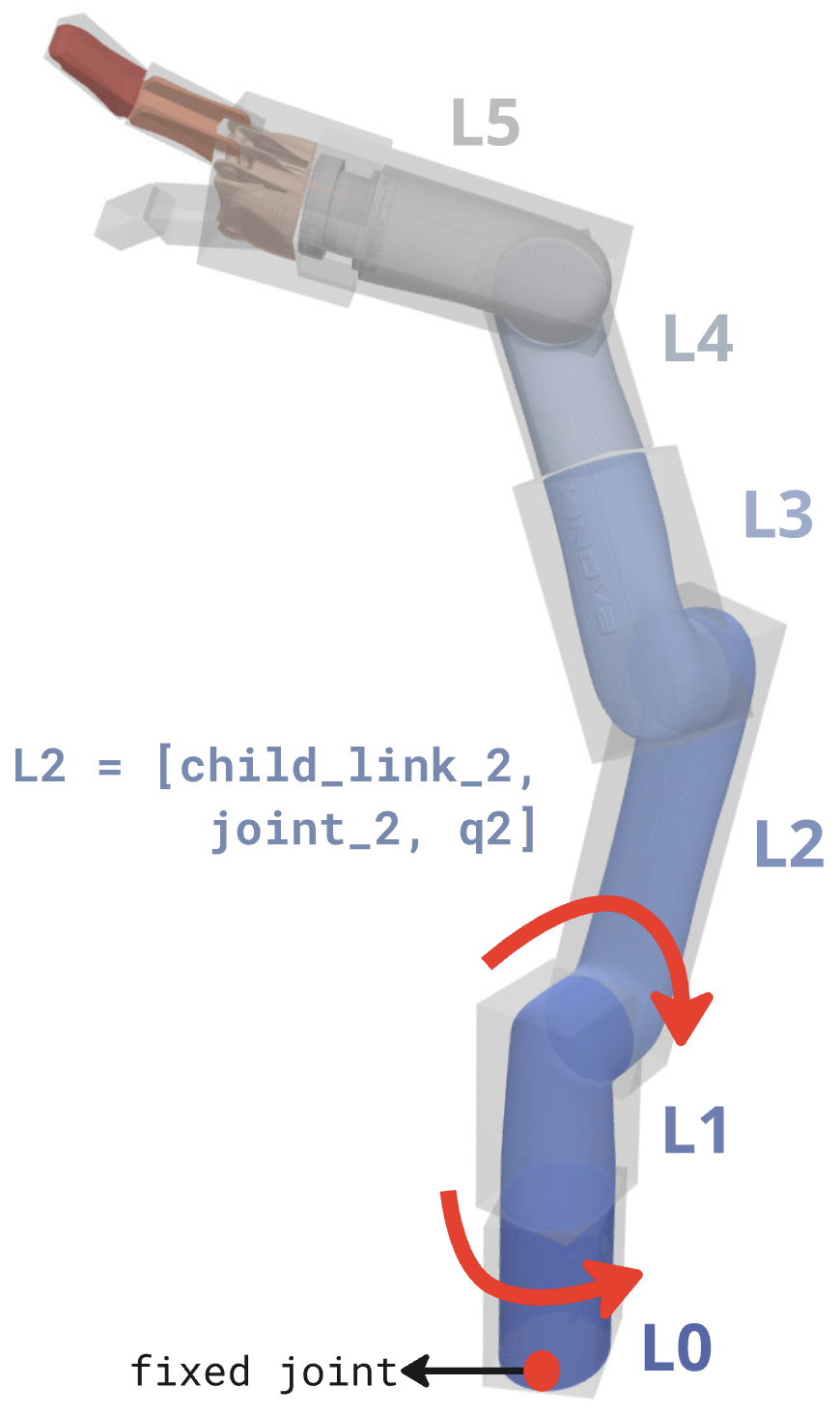}
  \caption{\footnotesize Robot morphology is represented by a sequence of links. We approximate the link geometries by the shape parameters of a fitted primitive.}
    \label{fig:link_geom}
\end{wrapfigure}

\textbf{Robot state.} The robot state is represented as a sequence of links, starting from the base link ($\mathbf{L}_0$) and followed by the subsequent robot links ($\mathbf{L}_i$). The state of each link \( i \) is described by its geometry information, the joint information (with which the link is attached to its parent), and the corresponding joint value \( q_i \) (see Figure \ref{fig:link_geom}). This can be expressed as:
\[
\mathbf{s_r} = \left[ \mathbf{L}_0, \mathbf{L}_1, \dots, \mathbf{L}_n \right]
\]

\vspace*{-15pt}
\textbf{Action Space.} 
We use joint-space control as the action space, where the policy outputs joint position changes $\Delta\mathbf{q}$. Control interfaces such as, end-effector control \cite{Yang2023PolybotTO} or unified action spaces \cite{Yang2024PushingTL}, are restrictive:(a) such a control interface ignores the collisions that may happen with the environment and robot body, (b) the end-effector link may not be the only link capable of causing good interaction with objects. An agent predicts $\Delta\mathbf{q}$ for all joints (including fixed joints), and we apply an action mask to select values for movable links only.

\textbf{Environment.}
For each of the benchmark tasks, the environments can optionally contain obstacles, which allows us to increase the difficulty of the task along the skill learning axis (learning a skill becomes harder in the presence of obstacles). For experiments, we only consider the obstacles variation for the \textit{Arm3} benchmark task. It is worth noting that the codebase is modular, allowing for the study of any combination of train-test robots, with or without obstacles, and the choice of observation space. 

See Appendix \ref{sec:env_modeling} for details on state space, rewards, and termination conditions.

\section{Baselines}
\begin{wrapfigure}{r}{0.2\textwidth}
  \centering
  \vspace*{-10pt}
    \includegraphics[width=0.2\textwidth]{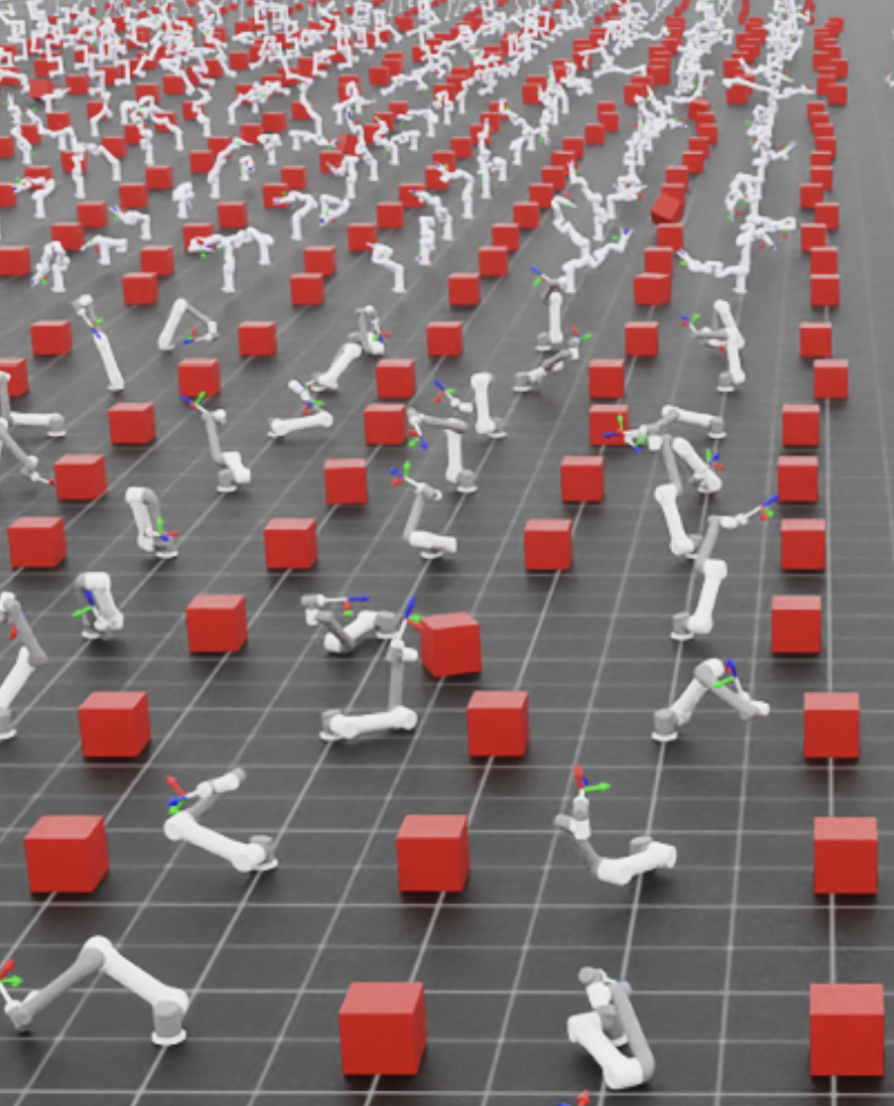}
      \vspace*{-15pt}
  \caption{\footnotesize Large-scale multi-embodiment training in Isaac-Sim.}
  \vspace*{-15pt}
  \label{fig:mt_isaac}
\end{wrapfigure}

\vspace*{-2pt}
\textbf{Setup} We evaluate RL agents on our benchmark suite, trained using the Isaac-Sim simulator Figure \ref{fig:mt_isaac}. Agents are trained with morphology-conditioned PPO \cite{Schulman2017ProximalPO} (and also goal-conditioned for \reach), aiming to maximize the average return across all training environments. 
We found PPO to be performing considerably better than other RL algorithms of SAC and TRPO (likely due to its lower sensitivity to hyperparameter tuning).
For each robot morphology, the return is the discounted sum of rewards, $R_i = \sum_{t=0}^{H}\gamma^t r_t^{(i)}$, where $r_t^{(i)}$ is the reward at time $t$ for robot $i$.

\textbf{Morphology-aware training.} We focus on two agent categories: single-embodiment (SE) and multi-embodiment (ME). SE agents are trained separately for each morphology using a dedicated actor-critic network. While they can't generalize to unseen morphologies, they serve as references to gauge the performance of ME agents. ME agents, in contrast, use a single actor-critic network trained across all training morphologies, enabling potential generalization. They collect experience from all train morphologies and compute combined policy and value loss for clipped updates, optimizing the average joint returns. To avoid any single morphology dominating the objective due to reward scale differences, we also experiment with task reweighting.  

\textbf{Policy architecture.} We evaluate two ME agent variations: MLP and Transformer \cite{Vaswani2017AttentionIA}. In both cases, each robot link and environment object is projected to a $d$-dimensional feature space. For the MLP, we flatten these features and output an action vector of size equal to the maximum number of links across embodiments. For the Transformer, we input the $d$-dimensional tokens, along with an observation mask (to ignore unavailable links), to the encoder. The encoder processes the sequence and outputs a feature vector for each link, which is then projected into scalar action values. These designs follow prior work \cite{Gupta2022MetaMorphLU}, and we refer to them as ME-MLP and ME-Tf. 

\subsection{Training Considerations}
\vspace*{-3pt}
While many areas remain open for improvement—such as extracting real robot link features or balancing learning across embodiments—we focus on two key challenges in the learning pipeline: improving training stability and addressing manipulation-specific needs.

\textbf{Symlog and critic updates.} For tasks such as \reach, the reward may depend on the end-effector's distance to the goal, the distribution of which varies across morphologies. Additionally, rewards increase sharply when the agent EE stays at the goal, causing reward scales to shift during training and destabilize critic learning. To address this, we adopt the \texttt{symlog} transformation from DreamerV3 \cite{dreamerv3}, to predict \texttt{symlog} of returns, and use its inverse (\texttt{symexp}) to compute target critic values. To further stabilize critic learning, we also experiment with slow critic updates, by doing an exponential moving average of weights.

\textbf{Discrete vs continuous actions.}
Training agents with continuous action outputs often leads to jittery motion and slower convergence. While reward engineering can help mitigate jitter, it is generally expensive and morphology dependent. Since we predict $\Delta\mathbf{q}$, we use discrete actions that simplify learning zero outputs. This also aligns with prior works that train on cross-embodiment data and discretize the action space \cite{kim2024openvla, brohan2022rt1}. The discrete variant of agents predicts logits over exponentially spaced bins around zero. 

\section{Experiments}
\vspace*{-3pt}

We design experiments to answer the following questions: (1) Can morphology-aware training outperform single-embodiment baselines? (2) Is zero-shot generalization to unseen morphologies feasible? (3) How do these trends vary across different tasks? and (4) How significantly do policy architectures and learning choices affect these outcomes? And finally, we examine the broader implications of our findings.

\subsection{Experiment Setup}
\textbf{RL Agents.} We train RL agents for both multi-embodiment (ME) and single-embodiment (SE) environments. In all experiments, we keep the same model size and training hyperparameters.  For \reach, we use the discrete version of the Transformer architecture. For \push, we use the default continuous version. We train all agents for at most 1M steps and report the best performance during training. See Appendix \ref{sec:training} for further details on RL training.

\textbf{Evaluation.} For \reach, the score is the average \textit{end-effector-to-goal} (EE-goal) reward over 10k evaluation steps, relative to a random agent. The EE-goal reward penalizes distance from the goal and rewards proximity. 
For \push, the score is the \textit{push success rate}: the proportion of episodes where the agent successfully pushes a block to the correct side, averaged over 10k evaluation steps. Please refer to the appendix for more details.
We report \textbf{Multi-task Score (MT)}, i.e., average score of all training morphologies, and  \textbf{Zero-shot Score (ZS)}, i.e., average score of unseen testing morphologies. 
The MT score reflects an agent's ability to optimize the learning objective, while the ZS score tests its generalization capability.

\subsection{Results}
\vspace{-5pt}

\begin{figure}[h!]
    \centering
    \begin{subfigure}[b]{0.49\textwidth}
        \includegraphics[width=\textwidth]{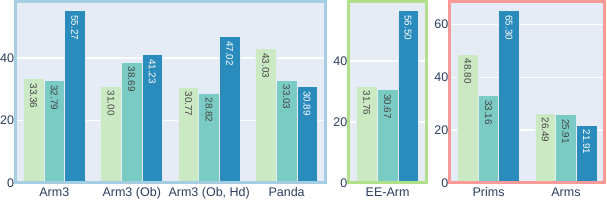} 
        \label{fig:subfig1}
    \end{subfigure}
    \hfill
    \begin{subfigure}[b]{0.49\textwidth}
        \includegraphics[trim=0 0 0 5, clip, width=\textwidth]{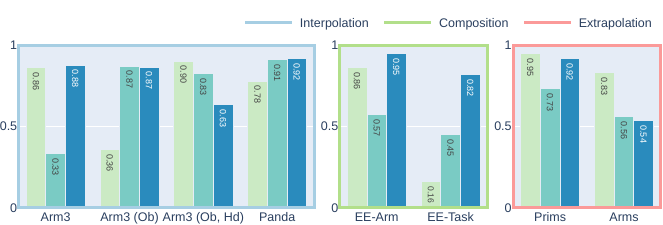}
        \label{fig:subfig2}
    \end{subfigure}
    \vspace*{-15pt}
    \caption{{\footnotesize MT score for (a) \reach, and (b) \push. The agents are: \colorbox[rgb]{0.8682814302191465, 0.9488811995386389, 0.8476585928489042}{\footnotesize Single-Embodiment}
\colorbox[rgb]{0.5847750865051903, 0.8386928104575164, 0.7344867358708189}{\footnotesize MLP}
\colorbox[rgb]{0.2084582852748943, 0.5934025374855825, 0.7689965397923876}{\footnotesize Transformer}
}}
    \label{fig:mt_plot}
    \vspace*{-5pt}
\end{figure}

\textbf{Q1. Can morphology-aware ME training outperform the SE baseline?} We focus on multi-task (MT) scores (Figure \ref{fig:mt_plot}) and observe that ME agents achieve comparable, or outperform, SE agents in most cases (except \textit{Panda} and \textit{Arms} experiments). When using the same backbone, the performance gain is small for interpolation. In others, ME performs worse. However, the Transformer-based ME agent consistently outperforms SE agents by a large margin in nearly all tasks.

\begin{wrapfigure}{r}{0.3\textwidth}
  \centering
    \includegraphics[width=0.28\textwidth]{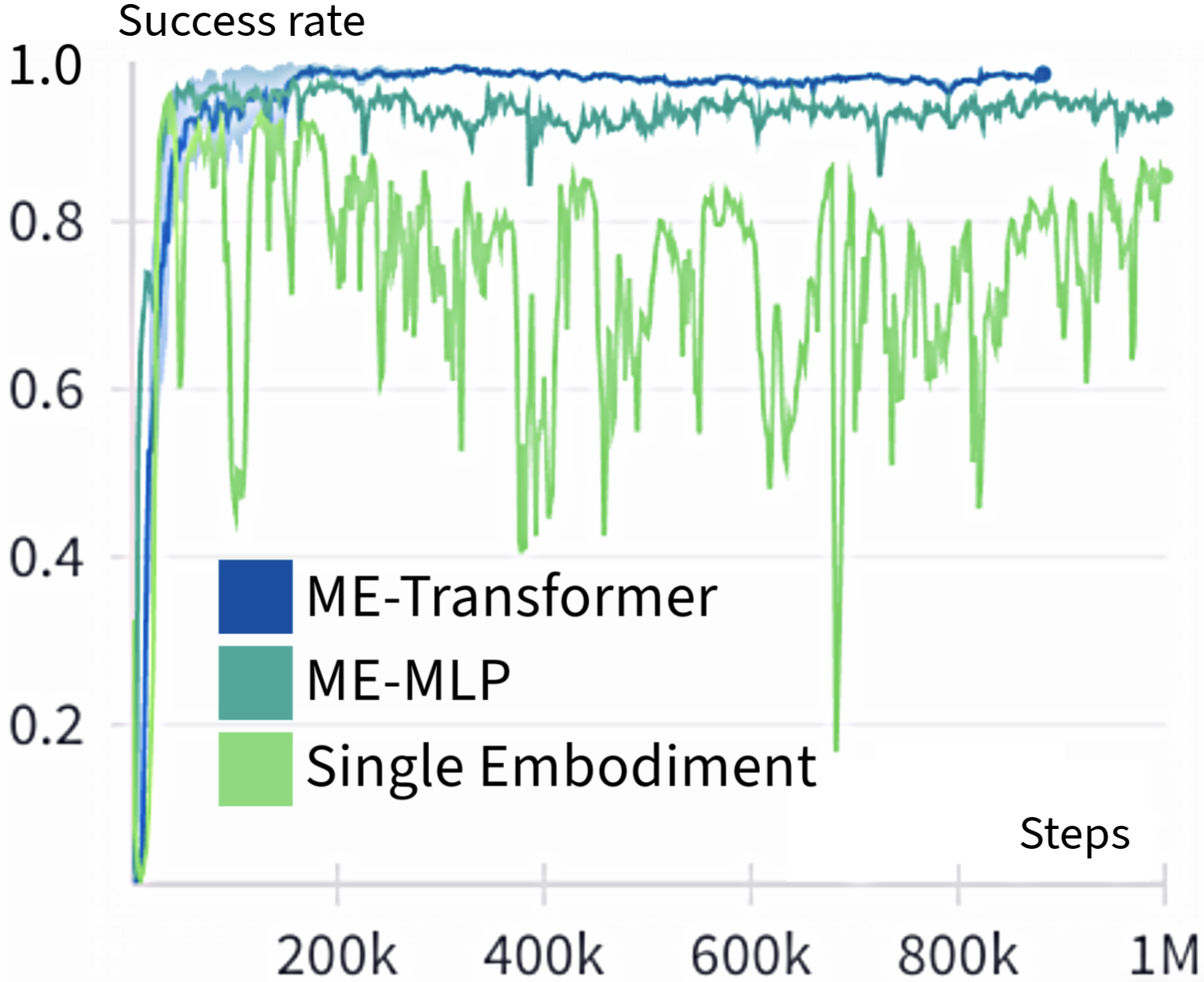}
  \caption{\footnotesize\textit{Panda}-\push~RL training with curriculum.}
  \vspace*{-15pt}
  \label{fig:regularization}
\end{wrapfigure}

We believe that ME training provides a regularizing effect, leading to more stable learning and reducing the chance of convergence to poor local minima. This can be seen in Figure \ref{fig:regularization} \textit{Panda}-\push~task, where ME agents show smoother learning curves. 
It is to note that unlike large-scale foundation model works that rely on extensive pretraining datasets, our setup controls for total environment interaction, isolating the impact of morphological diversity.

\begin{figure}
    \centering
    \begin{subfigure}[b]{0.49\textwidth}
        \includegraphics[width=\textwidth]{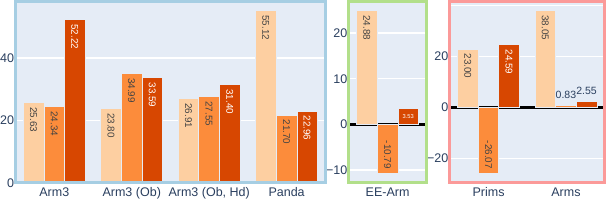} 
        \label{fig:subfig1}
    \end{subfigure}
    \hfill
    \begin{subfigure}[b]{0.49\textwidth}
        \includegraphics[trim=0 0 0 5, clip, width=\textwidth]{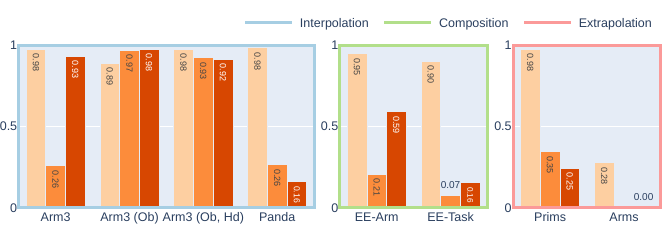} 
        \label{fig:subfig2}
    \end{subfigure}
    \vspace*{-15pt}
    \caption{{\footnotesize ZS score for (a) \reach, and (b) \push. The agents are: \colorbox[rgb]{0.9955709342560554, 0.8996539792387543, 0.7629988465974625}{\footnotesize Single-Embodiment}
\colorbox[rgb]{0.9905113417916186, 0.6576393694732795, 0.4468896578239139}{\footnotesize MLP}
\colorbox[rgb]{0.8704498269896194, 0.2485505574778931, 0.16822760476739718}{\footnotesize Transformer}
}}
    \label{fig:zs_plot}
    \vspace*{-15pt}
\end{figure}

\textbf{Q2. Can multi-robot training lead to better ZS generalization than training on a single robot?} Figure \ref{fig:zs_plot} shows that ME agents can outperform SE agents in zero-shot generalization, for \textit{Arm3} interpolation tasks, but have low success rates for the others.   

Current transformer-based agents used for morphology-aware learning lag far behind single-embodiment baselines on extrapolation and composition tasks, even when trained with $\sim$100M interactions. For instance, in the \textit{Arms}-\push~task, ME agents essentially fail with $0\%$ success rate. These results highlight the challenge of zero-shot generalization.

\textbf{Q3. How does agent performance compare across different tasks and benchmark categories?}
Results highlight that \textit{Arm3} is a simple environment, with multi-embodiment agents achieving high MT and ZS scores. In the same category, \textit{Panda} represents a more challenging task for both \reach~and \push, due to a more complex link structure. Composition tasks are harder than interpolation, with large performance gaps between SE and ME agents. Within composition, we observe that \textit{EE-Arm} is more challenging than \textit{EE-task}, because of the added inter-task complexity. In the extrapolation category, \textit{Prims} is an easier task than \textit{Arms}, though still challenging for \push; and \textit{Arms} is the most challenging for both \reach~and \push.

Across the manipulation tasks, \reach~is easier with simple morphologies (as higher complexity makes joint control challenging for precise goal reaching),  while \push~tasks seem less affected by complexity. Specifically, \textit{EE-Arm} struggles with \reach~but performs well in \push, while \textit{Prims} shows the opposite behavior.

\begin{wraptable}{r}{0.35\textwidth} 
\vspace{-10pt} 
  \centering
    \adjustbox{width=0.35\textwidth}{
    \begin{tabular}{l|c|c}
             \toprule
     Agent & Avg MT & ZS \\
     \midrule
 Rand& 0.0&0.0\\
Individual  &  33.36& 25.63\\
    \midrule
    Mlp&  32.79&  24.34\\
Tf&  8.83&  12.26\\
Tf + Sl&  32.84&  23.97\\
Tf + Sl + Dis&  \textbf{64.15}&  \textbf{61.91}\\
    Tf + Sl + Dis + TRW& 55.27& 52.22\\
\bottomrule
    \end{tabular}
    \vspace{-5pt}
    }
    \caption{\footnotesize Ablation study for \reach~in \textit{Arm3} interpolation environment. Mlp: MLP, Tf: transformer backbone, Sl: symlog returns, Dis: discrete actions, TRW: task re-weighting} 
    \label{tab:intra_bot_ablations}
    \vspace*{-0.5cm}
\end{wraptable}

\begin{wrapfigure}{r}{0.35\textwidth}
\vspace{-25pt}
  \centering
  \includegraphics[width=0.28\textwidth]{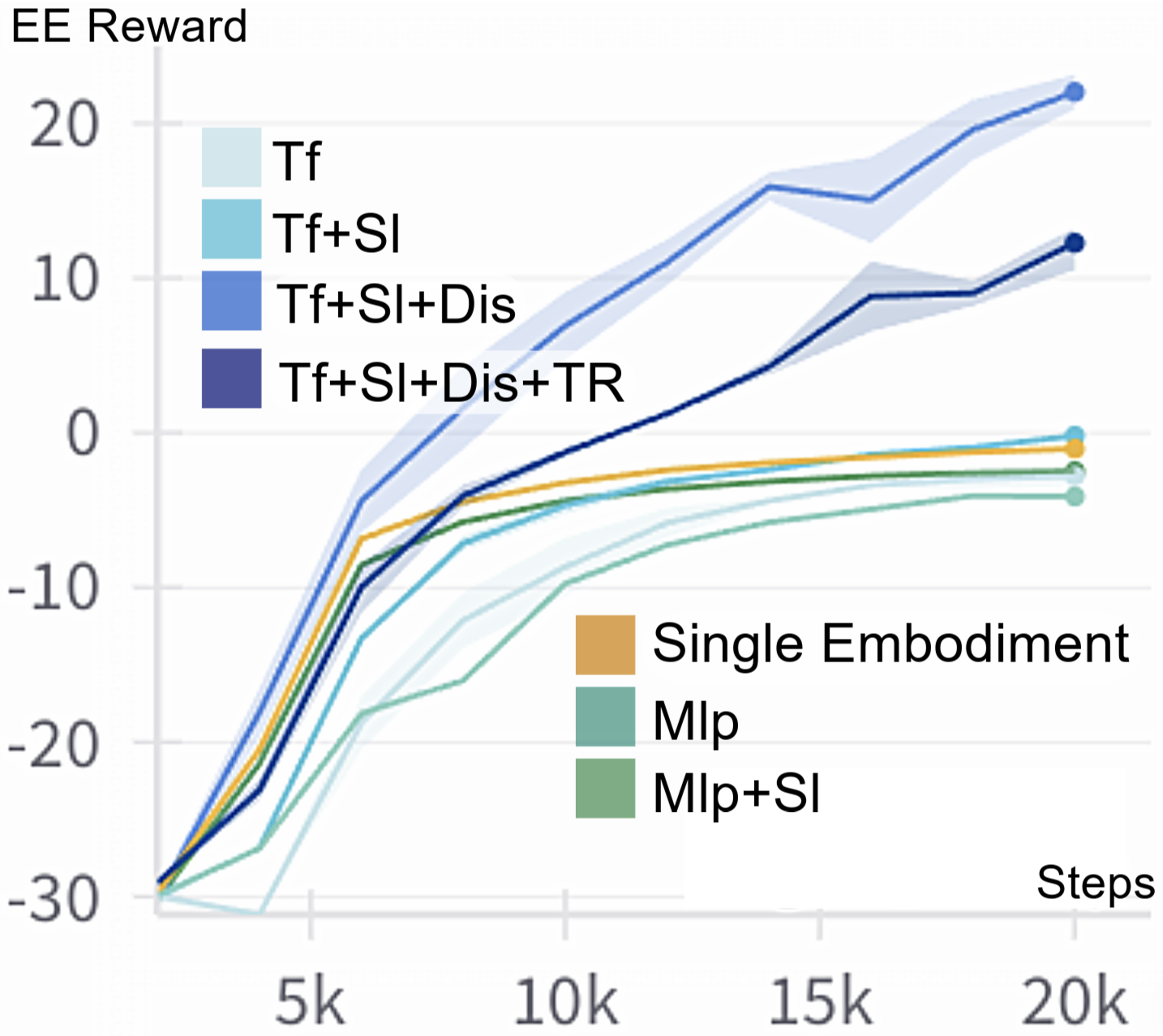} 
\caption{\footnotesize Ablation: Train curves for different variations. A discrete action space leads to faster convergence.}
\label{fig:convergence_plot}
\vspace*{-10pt}
\end{wrapfigure}

\textbf{Q4. How does architecture and policy design impact performance?} Table \ref{tab:intra_bot_ablations} shows that learning symlog of returns and using slow critic updates greatly benefits the ME-Tf agent, resulting in 4x and 2x performance for MT and ZS metrics, respectively. Additionally, training in a discrete action space is advantageous for the \reach~task, where precise goal-reaching is essential, leading to faster convergence (Figure \ref{fig:convergence_plot}). 

While random task re-weighting isn't necessary for tasks like \textit{Arm3}, we retain it to handle significant reward variations in other tasks, to prevent any single morphology from dominating the objective. Finally, the results for the \textit{Arm3} task demonstrate that combining \textit{both} Symlog and Discrete variation is crucial for ME-Tf agent to outperform SE agents. 

\textbf{Discussion.} The experiments highlight a key challenge: while Transformer agents match single-embodiment baselines in multi-task performance, they struggle with zero-shot generalization, especially in extrapolation and composition tasks. This indicates that optimizing for multi-embodiment performance alone cannot achieve strong generalization to out-of-distribution (OOD) morphologies. Consequently, as new robots with diverse morphologies enter the market, foundation models may struggle to adapt effectively. Our experiments demonstrate that simple multi-embodiment training fails to provide the crucial ability for agents to reason over morphologies and adapt to unseen robots. Further, the requirements for effective fine-tuning remain unclear.

In our fine-tuning experiments on test morphologies (see Appendix \ref{sec:finetuning}), we tested both 10k and 30k interaction steps. While fine-tuning shows significant improvements over training from scratch, even 30k steps of fine-tuning falls short compared to single-embodiment learning. This highlights the need for better multi-embodiment learning methods. Future research in this area should explore different policy architectures, better morphology representations, and applications of multi-task learning approaches to multi-embodiment learning (see Appendix \ref{sec:future} for further discussion).

\vspace*{5pt}
\textbf{Conclusion.} We propose \bench, a benchmark suite for evaluating morphology generalization across three key axes: interpolation, extrapolation, and composition. Our experiments on multi-embodiment RL agents show that while multi-embodiment training improves in-distribution performance, it struggles with zero-shot generalization to novel morphology structures. Our experiments also demonstrate the importance of design choices like action space representation and learning stability techniques in achieving good performance. This work opens possibilities for future research to build robotic systems capable of true morphological generalization. 

\section*{Limitations}

First, we have focused on controlling and understanding the use of diverse embodiments in this paper, and have chosen to largely ignore the visual variation in benchmark tasks. However, the use of visual variations, and RGB inputs is a common setting of works in robotic manipulation,  including methods such as \cite{Hu2021KnowTT}, or \cite{yuan2024learning}. We consider these types of methods as complementary to our benchmark, as they focus on how to align vision inputs, while we assume that we have already separated embodiment from environment information. Further, training policies without visual inputs is faster and less compute intensive, and almost complementary to advances in computer vision. 

Second, while we aim to bring RL benchmarks close to real robot tasks, our tasks are still quite simple and basic. Our ongoing work aims to add more tasks to the benchmark: articulate object manipulation, including doors, drawers, and knobs. 

Finally, training RL policies for the different variations and different benchmark tasks requires a pretty significant GPU training. We hope that future methods can focus on efficient adaptation methods to improve learning. 

\section*{Acknowledgment}
This work was partially supported by the National Science Foundation. 

\bibliography{main}

\begin{thebibliography}{43}
\providecommand{\natexlab}[1]{#1}
\providecommand{\url}[1]{\texttt{#1}}
\expandafter\ifx\csname urlstyle\endcsname\relax
  \providecommand{\doi}[1]{doi: #1}\else
  \providecommand{\doi}{doi: \begingroup \urlstyle{rm}\Url}\fi

\bibitem[et. al.(2024{\natexlab{a}})]{ONeill2024OpenXR}
A.~O. et. al.
\newblock Open x-embodiment: Robotic learning datasets and rt-x models : Open x-embodiment collaboration0.
\newblock \emph{2024 IEEE International Conference on Robotics and Automation (ICRA)}, pages 6892--6903, 2024{\natexlab{a}}.
\newblock URL \url{https://arxiv.org/abs/2310.08864}.

\bibitem[et. al.(2024{\natexlab{b}})]{Khazatsky2024DROIDAL}
A.~K. et. al.
\newblock Droid: A large-scale in-the-wild robot manipulation dataset.
\newblock \emph{ArXiv}, abs/2403.12945, 2024{\natexlab{b}}.

\bibitem[Gupta et~al.(2022)Gupta, Fan, Ganguli, and Fei-Fei]{Gupta2022MetaMorphLU}
A.~Gupta, L.~J. Fan, S.~Ganguli, and L.~Fei-Fei.
\newblock Metamorph: Learning universal controllers with transformers.
\newblock \emph{ArXiv}, abs/2203.11931, 2022.

\bibitem[Patel and Song(2024)]{patel2024getzero}
A.~Patel and S.~Song.
\newblock {GET-Zero}: Graph embodiment transformer for zero-shot embodiment generalization.
\newblock \emph{2025 IEEE International Conference on Robotics and Automation (ICRA)}, 2024.
\newblock URL \url{https://arxiv.org/abs/2407.15002}.

\bibitem[Yang et~al.(2024)Yang, Glossop, Bhorkar, Shah, Vuong, Finn, Sadigh, and Levine]{Yang2024PushingTL}
J.~Yang, C.~Glossop, A.~Bhorkar, D.~Shah, Q.~Vuong, C.~Finn, D.~Sadigh, and S.~Levine.
\newblock Pushing the limits of cross-embodiment learning for manipulation and navigation.
\newblock \emph{ArXiv}, abs/2402.19432, 2024.

\bibitem[Kim et~al.(2024)Kim, Pertsch, Karamcheti, Xiao, Balakrishna, Nair, Rafailov, Foster, Lam, Sanketi, Vuong, Kollar, Burchfiel, Tedrake, Sadigh, Levine, Liang, and Finn]{Kim2024OpenVLAAO}
M.~J. Kim, K.~Pertsch, S.~Karamcheti, T.~Xiao, A.~Balakrishna, S.~Nair, R.~Rafailov, E.~Foster, G.~Lam, P.~R. Sanketi, Q.~Vuong, T.~Kollar, B.~Burchfiel, R.~Tedrake, D.~Sadigh, S.~Levine, P.~Liang, and C.~Finn.
\newblock Openvla: An open-source vision-language-action model.
\newblock \emph{ArXiv}, abs/2406.09246, 2024.

\bibitem[Team et~al.(2024)Team, Ghosh, Walke, Pertsch, Black, Mees, Dasari, Hejna, Kreiman, Xu, Luo, Tan, Sanketi, Vuong, Xiao, Sadigh, Finn, and Levine]{Team2024OctoAO}
O.~M. Team, D.~Ghosh, H.~Walke, K.~Pertsch, K.~Black, O.~Mees, S.~Dasari, J.~Hejna, T.~Kreiman, C.~Xu, J.~Luo, Y.~L. Tan, P.~R. Sanketi, Q.~Vuong, T.~Xiao, D.~Sadigh, C.~Finn, and S.~Levine.
\newblock Octo: An open-source generalist robot policy.
\newblock \emph{ArXiv}, abs/2405.12213, 2024.

\bibitem[Chen et~al.(2018)Chen, Murali, and Gupta]{Chen2018HardwareCP}
T.~Chen, A.~Murali, and A.~K. Gupta.
\newblock Hardware conditioned policies for multi-robot transfer learning.
\newblock In \emph{Neural Information Processing Systems}, 2018.

\bibitem[Hu et~al.(2021)Hu, Huang, Rybkin, and Jayaraman]{Hu2021KnowTT}
E.~S. Hu, K.-Y. Huang, O.~Rybkin, and D.~Jayaraman.
\newblock Know thyself: Transferable visuomotor control through robot-awareness.
\newblock \emph{ArXiv}, abs/2107.09047, 2021.

\bibitem[Sferrazza et~al.(2024)Sferrazza, Huang, Liu, Lee, and Abbeel]{sferrazza2024body}
C.~Sferrazza, D.-M. Huang, F.~Liu, J.~Lee, and P.~Abbeel.
\newblock Body transformer: Leveraging robot embodiment for policy learning.
\newblock \emph{arXiv preprint arXiv:2408.06316}, 2024.

\bibitem[Kurin et~al.(2021)Kurin, Igl, Rockt{\"a}schel, Boehmer, and Whiteson]{kurin2021my}
V.~Kurin, M.~Igl, T.~Rockt{\"a}schel, W.~Boehmer, and S.~Whiteson.
\newblock My body is a cage: the role of morphology in graph-based incompatible control.
\newblock In \emph{International Conference on Learning Representations}, 2021.
\newblock URL \url{https://openreview.net/forum?id=N3zUDGN5lO}.

\bibitem[Wang et~al.(2018)Wang, Liao, Ba, and Fidler]{Wang2018NerveNetLS}
T.~Wang, R.~Liao, J.~Ba, and S.~Fidler.
\newblock Nervenet: Learning structured policy with graph neural networks.
\newblock In \emph{International Conference on Learning Representations}, 2018.

\bibitem[Yang et~al.(2023)Yang, Sadigh, and Finn]{Yang2023PolybotTO}
J.~Yang, D.~Sadigh, and C.~Finn.
\newblock Polybot: Training one policy across robots while embracing variability.
\newblock In \emph{Conference on Robot Learning}, 2023.

\bibitem[Yu et~al.(2021)Yu, Quillen, He, Julian, Narayan, Shively, Bellathur, Hausman, Finn, and Levine]{yu2021metaworldbenchmarkevaluationmultitask}
T.~Yu, D.~Quillen, Z.~He, R.~Julian, A.~Narayan, H.~Shively, A.~Bellathur, K.~Hausman, C.~Finn, and S.~Levine.
\newblock Meta-world: A benchmark and evaluation for multi-task and meta reinforcement learning, 2021.
\newblock URL \url{https://arxiv.org/abs/1910.10897}.

\bibitem[{NVIDIA Corporation}(2024)]{isaacsim}
{NVIDIA Corporation}.
\newblock Nvidia isaac sim, 2024.
\newblock Version 4.0. Available at \url{https://developer.nvidia.com/isaac/sim}.

\bibitem[Mittal et~al.(2023)Mittal, Yu, Yu, Liu, Rudin, Hoeller, Yuan, Singh, Guo, Mazhar, Mandlekar, Babich, State, Hutter, and Garg]{isaaclab}
M.~Mittal, C.~Yu, Q.~Yu, J.~Liu, N.~Rudin, D.~Hoeller, J.~L. Yuan, R.~Singh, Y.~Guo, H.~Mazhar, A.~Mandlekar, B.~Babich, G.~State, M.~Hutter, and A.~Garg.
\newblock Orbit: A unified simulation framework for interactive robot learning environments.
\newblock \emph{IEEE Robotics and Automation Letters}, 8\penalty0 (6):\penalty0 3740--3747, 2023.
\newblock \doi{10.1109/LRA.2023.3270034}.

\bibitem[Makoviychuk et~al.(2022)]{isaaclab2022}
V.~Makoviychuk et~al.
\newblock Isaac lab: A unified framework for robot learning.
\newblock \url{https://github.com/isaac-sim/IsaacLab}, 2022.

\bibitem[Bjorck et~al.(2025)Bjorck, Casta{\~n}eda, Cherniadev, Da, Ding, Fan, Fang, Fox, Hu, Huang, et~al.]{bjorck2025gr00t}
J.~Bjorck, F.~Casta{\~n}eda, N.~Cherniadev, X.~Da, R.~Ding, L.~Fan, Y.~Fang, D.~Fox, F.~Hu, S.~Huang, et~al.
\newblock Gr00t n1: An open foundation model for generalist humanoid robots.
\newblock \emph{arXiv preprint arXiv:2503.14734}, 2025.

\bibitem[Black et~al.(2024)Black, Brown, Driess, Esmail, Equi, Finn, Fusai, Groom, Hausman, Ichter, Jakubczak, Jones, Ke, Levine, Li-Bell, Mothukuri, Nair, Pertsch, Shi, Tanner, Vuong, Walling, Wang, and Zhilinsky]{black2024pi0}
K.~Black, N.~Brown, D.~Driess, A.~Esmail, M.~Equi, C.~Finn, N.~Fusai, L.~Groom, K.~Hausman, B.~Ichter, S.~Jakubczak, T.~Jones, L.~Ke, S.~Levine, A.~Li-Bell, M.~Mothukuri, S.~Nair, K.~Pertsch, L.~X. Shi, J.~Tanner, Q.~Vuong, A.~Walling, H.~Wang, and U.~Zhilinsky.
\newblock $\pi_0$: A vision-language-action flow model for general robot control, 2024.
\newblock URL \url{https://arxiv.org/abs/2410.24164}.

\bibitem[Kim et~al.(2024)Kim, Pertsch, Karamcheti, Xiao, Balakrishna, Nair, Rafailov, Foster, Lam, Sanketi, et~al.]{kim2024openvla}
M.~J. Kim, K.~Pertsch, S.~Karamcheti, T.~Xiao, A.~Balakrishna, S.~Nair, R.~Rafailov, E.~Foster, G.~Lam, P.~Sanketi, et~al.
\newblock Openvla: An open-source vision-language-action model.
\newblock \emph{arXiv preprint arXiv:2406.09246}, 2024.

\bibitem[Team et~al.(2024)Team, Ghosh, Walke, Pertsch, Black, Mees, Dasari, Hejna, Kreiman, Xu, et~al.]{team2024octo}
O.~M. Team, D.~Ghosh, H.~Walke, K.~Pertsch, K.~Black, O.~Mees, S.~Dasari, J.~Hejna, T.~Kreiman, C.~Xu, et~al.
\newblock Octo: An open-source generalist robot policy.
\newblock \emph{arXiv preprint arXiv:2405.12213}, 2024.

\bibitem[Bousmalis et~al.(2023)Bousmalis, Vezzani, Rao, Devin, Lee, Bauz{\'a}, Davchev, Zhou, Gupta, Raju, et~al.]{bousmalis2023robocat}
K.~Bousmalis, G.~Vezzani, D.~Rao, C.~Devin, A.~X. Lee, M.~Bauz{\'a}, T.~Davchev, Y.~Zhou, A.~Gupta, A.~Raju, et~al.
\newblock Robocat: A self-improving generalist agent for robotic manipulation.
\newblock \emph{arXiv preprint arXiv:2306.11706}, 2023.

\bibitem[Reed et~al.(2022)Reed, Zolna, Parisotto, Colmenarejo, Novikov, Barth-Maron, Gim{\'e}nez, Sulsky, Kay, Springenberg, Eccles, Bruce, Razavi, Edwards, Heess, Chen, Hadsell, Vinyals, Bordbar, and de~Freitas]{Reed2022AGA}
S.~Reed, K.~Zolna, E.~Parisotto, S.~G. Colmenarejo, A.~Novikov, G.~Barth-Maron, M.~Gim{\'e}nez, Y.~Sulsky, J.~Kay, J.~T. Springenberg, T.~Eccles, J.~Bruce, A.~Razavi, A.~D. Edwards, N.~M.~O. Heess, Y.~Chen, R.~Hadsell, O.~Vinyals, M.~Bordbar, and N.~de~Freitas.
\newblock A generalist agent.
\newblock \emph{ArXiv}, abs/2205.06175, 2022.

\bibitem[Gu et~al.(2023)Gu, Xiang, Li, Ling, Liu, Mu, Tang, Tao, Wei, Yao, et~al.]{gu2023maniskill2}
J.~Gu, F.~Xiang, X.~Li, Z.~Ling, X.~Liu, T.~Mu, Y.~Tang, S.~Tao, X.~Wei, Y.~Yao, et~al.
\newblock Maniskill2: A unified benchmark for generalizable manipulation skills.
\newblock \emph{arXiv preprint arXiv:2302.04659}, 2023.

\bibitem[Ehsani et~al.(2021)Ehsani, Han, Herrasti, VanderBilt, Weihs, Kolve, Kembhavi, and Mottaghi]{ehsani2021manipulathor}
K.~Ehsani, W.~Han, A.~Herrasti, E.~VanderBilt, L.~Weihs, E.~Kolve, A.~Kembhavi, and R.~Mottaghi.
\newblock Manipulathor: A framework for visual object manipulation.
\newblock In \emph{Proceedings of the IEEE/CVF conference on computer vision and pattern recognition}, pages 4497--4506, 2021.

\bibitem[James et~al.(2020)James, Ma, Arrojo, and Davison]{james2020rlbench}
S.~James, Z.~Ma, D.~R. Arrojo, and A.~J. Davison.
\newblock Rlbench: The robot learning benchmark \& learning environment.
\newblock \emph{IEEE Robotics and Automation Letters}, 5\penalty0 (2):\penalty0 3019--3026, 2020.

\bibitem[Zhu et~al.(2020)Zhu, Wong, Mandlekar, Mart{\'\i}n-Mart{\'\i}n, Joshi, Nasiriany, and Zhu]{zhu2020robosuite}
Y.~Zhu, J.~Wong, A.~Mandlekar, R.~Mart{\'\i}n-Mart{\'\i}n, A.~Joshi, S.~Nasiriany, and Y.~Zhu.
\newblock robosuite: A modular simulation framework and benchmark for robot learning.
\newblock \emph{arXiv preprint arXiv:2009.12293}, 2020.

\bibitem[Yu et~al.(2020)Yu, Quillen, He, Julian, Hausman, Finn, and Levine]{yu2020meta}
T.~Yu, D.~Quillen, Z.~He, R.~Julian, K.~Hausman, C.~Finn, and S.~Levine.
\newblock Meta-world: A benchmark and evaluation for multi-task and meta reinforcement learning.
\newblock In \emph{Conference on robot learning}, pages 1094--1100. PMLR, 2020.

\bibitem[Heo et~al.(2023)Heo, Lee, Lee, and Lim]{heo2023furniturebench}
M.~Heo, Y.~Lee, D.~Lee, and J.~J. Lim.
\newblock Furniturebench: Reproducible real-world benchmark for long-horizon complex manipulation.
\newblock In \emph{Robotics: Science and Systems}, 2023.

\bibitem[Han et~al.(2024)Han, Parakh, Geng, Defay, Luyang, and Deng]{han2024fetchbench}
B.~Han, M.~Parakh, D.~Geng, J.~A. Defay, G.~Luyang, and J.~Deng.
\newblock Fetchbench: A simulation benchmark for robot fetching.
\newblock \emph{arXiv preprint arXiv:2406.11793}, 2024.

\bibitem[Li et~al.(2023)Li, Zhang, Wong, Gokmen, Srivastava, Mart{\'\i}n-Mart{\'\i}n, Wang, Levine, Lingelbach, Sun, et~al.]{li2023behavior1k}
C.~Li, R.~Zhang, J.~Wong, C.~Gokmen, S.~Srivastava, R.~Mart{\'\i}n-Mart{\'\i}n, C.~Wang, G.~Levine, M.~Lingelbach, J.~Sun, et~al.
\newblock Behavior-1k: A benchmark for embodied ai with 1,000 everyday activities and realistic simulation.
\newblock In \emph{Conference on Robot Learning}, pages 80--93. PMLR, 2023.

\bibitem[Srivastava et~al.(2022)Srivastava, Li, Lingelbach, Mart{\'\i}n-Mart{\'\i}n, Xia, Vainio, Lian, Gokmen, Buch, Liu, et~al.]{srivastava2022behavior100}
S.~Srivastava, C.~Li, M.~Lingelbach, R.~Mart{\'\i}n-Mart{\'\i}n, F.~Xia, K.~E. Vainio, Z.~Lian, C.~Gokmen, S.~Buch, K.~Liu, et~al.
\newblock Behavior: Benchmark for everyday household activities in virtual, interactive, and ecological environments.
\newblock In \emph{Conference on robot learning}, pages 477--490. PMLR, 2022.

\bibitem[Szot et~al.(2021)Szot, Clegg, Undersander, Wijmans, Zhao, Turner, Maestre, Mukadam, Chaplot, Maksymets, et~al.]{szot2021habitat2}
A.~Szot, A.~Clegg, E.~Undersander, E.~Wijmans, Y.~Zhao, J.~Turner, N.~Maestre, M.~Mukadam, D.~S. Chaplot, O.~Maksymets, et~al.
\newblock Habitat 2.0: Training home assistants to rearrange their habitat.
\newblock \emph{Advances in neural information processing systems}, 34:\penalty0 251--266, 2021.

\bibitem[Sferrazza et~al.(2024)Sferrazza, Huang, Lin, Lee, and Abbeel]{sferrazza2024humanoidbench}
C.~Sferrazza, D.-M. Huang, X.~Lin, Y.~Lee, and P.~Abbeel.
\newblock Humanoidbench: Simulated humanoid benchmark for whole-body locomotion and manipulation, 2024.

\bibitem[Gupta et~al.(2021)Gupta, Savarese, Ganguli, and Fei-Fei]{unimal}
A.~Gupta, S.~Savarese, S.~Ganguli, and L.~Fei-Fei.
\newblock Embodied intelligence via learning and evolution.
\newblock \emph{Nature Communications}, 12\penalty0 (1), Oct. 2021.
\newblock ISSN 2041-1723.
\newblock \doi{10.1038/s41467-021-25874-z}.
\newblock URL \url{http://dx.doi.org/10.1038/s41467-021-25874-z}.

\bibitem[Hong et~al.(2022)Hong, Yoon, and Kim]{Hong2022StructureAwareTP}
S.~Hong, D.~Yoon, and K.-E. Kim.
\newblock Structure-aware transformer policy for inhomogeneous multi-task reinforcement learning.
\newblock In \emph{International Conference on Learning Representations}, 2022.

\bibitem[Zhang et~al.(2020)Zhang, Xiao, Efros, Pinto, and Wang]{zhang2020crossloco}
Q.~Zhang, T.~Xiao, A.~A. Efros, L.~Pinto, and X.~Wang.
\newblock Learning cross-domain correspondence for control with dynamics cycle-consistency.
\newblock \emph{arXiv preprint arXiv:2012.09811}, 2020.

\bibitem[Schulman et~al.(2017)Schulman, Wolski, Dhariwal, Radford, and Klimov]{Schulman2017ProximalPO}
J.~Schulman, F.~Wolski, P.~Dhariwal, A.~Radford, and O.~Klimov.
\newblock Proximal policy optimization algorithms.
\newblock \emph{ArXiv}, abs/1707.06347, 2017.

\bibitem[Vaswani et~al.(2017)Vaswani, Shazeer, Parmar, Uszkoreit, Jones, Gomez, Kaiser, and Polosukhin]{Vaswani2017AttentionIA}
A.~Vaswani, N.~M. Shazeer, N.~Parmar, J.~Uszkoreit, L.~Jones, A.~N. Gomez, L.~Kaiser, and I.~Polosukhin.
\newblock Attention is all you need.
\newblock In \emph{Neural Information Processing Systems}, 2017.

\bibitem[Hafner et~al.(2024)Hafner, Pasukonis, Ba, and Lillicrap]{dreamerv3}
D.~Hafner, J.~Pasukonis, J.~Ba, and T.~Lillicrap.
\newblock Mastering diverse domains through world models, 2024.
\newblock URL \url{https://arxiv.org/abs/2301.04104}.

\bibitem[Brohan et~al.(2022)Brohan, Brown, Carbajal, Chebotar, Dabis, Finn, Gopalakrishnan, Hausman, Herzog, Hsu, et~al.]{brohan2022rt1}
A.~Brohan, N.~Brown, J.~Carbajal, Y.~Chebotar, J.~Dabis, C.~Finn, K.~Gopalakrishnan, K.~Hausman, A.~Herzog, J.~Hsu, et~al.
\newblock Rt-1: Robotics transformer for real-world control at scale.
\newblock \emph{arXiv preprint arXiv:2212.06817}, 2022.

\bibitem[Yuan et~al.(2024)Yuan, Wei, Cheng, Zhang, Chen, and Xu]{yuan2024learning}
Z.~Yuan, T.~Wei, S.~Cheng, G.~Zhang, Y.~Chen, and H.~Xu.
\newblock Learning to manipulate anywhere: A visual generalizable framework for reinforcement learning.
\newblock In \emph{8th Annual Conference on Robot Learning}, 2024.
\newblock URL \url{https://openreview.net/forum?id=jart4nhCQr}.

\bibitem[Serrano-Muñoz et~al.(2023)Serrano-Muñoz, Chrysostomou, Bøgh, and Arana-Arexolaleiba]{serrano2023skrl}
A.~Serrano-Muñoz, D.~Chrysostomou, S.~Bøgh, and N.~Arana-Arexolaleiba.
\newblock skrl: Modular and flexible library for reinforcement learning.
\newblock \emph{Journal of Machine Learning Research}, 24\penalty0 (254):\penalty0 1--9, 2023.
\newblock URL \url{http://jmlr.org/papers/v24/23-0112.html}.

\end{thebibliography}

\newpage

\section*{\Large Appendix}
\setcounter{section}{0}

\section{Benchmark Details}\label{sec:benchmark_section}

\subsection{Benchmark Tasks}\label{sec:benchmark_stats}
Table \ref{tab:bench_tasks_statistics} shows the statistics for train morphologies in the benchmark tasks. The number of movable joints (till EE) is more relevant for \reach, where we are controlling to make EE reach the goal, and does not aim to control joints beyond EE, for example, finger joints. 
\vspace{-5pt}
\begin{table}[h]
\centering
    \caption{\small Statistics about benchmark tasks.}
    \label{tab:bench_tasks_statistics}
    \vspace{2pt}
\resizebox{\columnwidth}{!}{
        \begin{tabular}{l|l|c|c|c|c}
        \toprule
        \multicolumn{2}{c|}{Benchmark Tasks} & Avg \# links& Avg \# movable joints &\# movable joints (till EE) & Cosine similarity \\ 
        \midrule
        \rowcolor{cyan!4} Interpolation& Arm-3 & 10& 9& 4& 0.80\\
        \rowcolor{cyan!4}& Panda & 11& 9& 7& 0.76\\
         \midrule
         \rowcolor{green!4} Composition& EE Arm & 8.5& 5.9& 5.6 & 0.47\\
        \rowcolor{green!4}& EE-Task  & 7.75& 5.25& - & 0.49\\
        \midrule
        \rowcolor{red!3} Extrapolation& Primitives & 6.60& 4.60& 4.60 &0.45\\
        \rowcolor{red!3}& Robot Arms & 14.70& 8.5& 6.6& 0.42\\
        \bottomrule
        \end{tabular}
        }
\end{table}

\subsection{Manipulation Tasks}\label{sec:manip_tasks}
\paragraph{\reach~task} In reach environments, the objective is to move the EE to a given goal position by controlling the joint positions while avoiding obstacles. For each benchmark task, we generate 100 end-effector goals per robot by randomly sampling joint positions that result in non-colliding configurations and computing the corresponding EE pose. At the start of each episode during training, a goal is uniformly selected from the pre-generated EE poses. 

\paragraph{\push~task} The block starts at a position with $y < -15.0 cm$ and must be pushed towards the positive side of the y-axis until its center crosses $y = 20 cm$. 

\section{Environment Modeling}\label{sec:env_modeling}
\subsection{State space}\label{sec:state_space}

\paragraph{Robot State} Each link in the robot state is represented by a 48-dimensional vector, with its components detailed in Table \ref{tab:robo_link_state}.

\begin{table}[h]
    \centering
    \caption{\small Each link is encoded as a 48-dimensional vector, capturing information about the link, its joint, and joint value.}
    \label{tab:robo_link_state}
    \vspace{5pt}
    \begin{tabular}{lcp{8cm}}
        \toprule
        Link Information & Dimension & Description \\
        \midrule
        Link Index & 1 & Index of the given link. \\
        Parent Index & 1 & Index of the parent link. \\
        EE flag & 4 & One-hot encoded indicator for whether the link is an end-effector (repeated 4x). \\
        Geometry type & 6 & One-hot encoding of the shape type (3 dimensions for different types) \& 3 dimensions for shape parameters along the [x, y, z] axes. \\
        Link origin & 7 & 3D position [x, y, z] and orientation as a quaternion. \\
        \midrule
        Joint axis & 3 & Unit vector representing the axis of rotation or translation of the joint. \\
        Joint origin & 7 & 3D position [x, y, z] and orientation as a quaternion. \\
        Joint type & 3 & One-hot encoded vector indicating the type of joint: [prismatic, revolute, fixed]. \\
        \midrule
        Joint pos (\(q\)) & 16 & Sinusoidal encoding of the joint value \(q\). \\
        \midrule
        \textbf{Total} & 48 & -\\
        \bottomrule
    \end{tabular}
\end{table}

\paragraph{Link Geometry} For artificial robots, the geometry is directly encoded as they are made of primitive shapes. For real robots, we use the Trimesh library to obtain the best-fit box, sphere, and cylinder, selecting the one with the lowest fit error (volume overlap).

\subsection{Reward}\label{sec:reward}
The total reward for \reach~is computed as a weighted sum of four components:
\[
R = w_1 r_{\text{joint-limits}} + w_2 r_{\text{joint-acc}} + w_3 r_{\text{ee-goal}} + w_4 r_{\text{vicinity}},
\]
where \( w_i \) are scalar weights corresponding to each reward term. These reward terms are detailed in Table \ref{tab:reward_terms_reach}.

The total reward for \push~is a weighted sum of three components:
\[
R = w_1 r_{\text{joint-limits}} + w_2 r_{\text{obj-dist}} + w_3 r_{\text{termination}},
\]
where \( w_i \) are scalar weights for each reward term. The reward terms are described in Table \ref{tab:reward_terms_push}.

\begin{table}[h]
    \centering
    \caption{\small Reward terms for \reach~task.}
    \label{tab:reward_terms_reach}
    \begin{tabular}{lp{10cm}}
        \toprule
        Reward Term & Expression \\
       \midrule
        \( r_{\text{joint-limits}} \) & \( r_{\text{joint-limits}} = -\max(0, q - q_{\text{upper}}, q_{\text{lower}} - q) \) \\[5pt]
        & \small This term penalizes the robot's joint positions \( q \) if they violate predefined soft upper (\( q_{\text{upper}} \)) or lower (\( q_{\text{lower}} \)) limits. The penalty increases with the extent of the violation. \\
               \midrule
        \( r_{\text{joint-acc}} \) & \( r_{\text{joint-acc}} = -\ddot{q} \) \\[5pt]
        & \small This term discourages high joint accelerations \( \ddot{q} \), promoting smoother and more energy-efficient movements. \\
               \midrule
        \( r_{\text{ee-goal}} \) & \( r_{\text{ee-goal}} = -\text{dist}(\text{EE}, \text{goal}) \) \\[5pt]
        & \small This term provides a continuous reward inversely proportional to the Euclidean distance between the robot's EE and the target goal position. Minimizing this distance maximizes the reward. \\
               \midrule
        \( r_{\text{vicinity}} \) & \( r_{\text{vicinity}} = \begin{cases} 1.0, & \text{if } \text{dist}(\text{EE}, \text{goal}) < \text{threshold} \\ 0, & \text{otherwise} \end{cases} \) \\[8pt]
        & \small This term offers a discrete, positive reward of 1.0 when the end-effector is within a specified distance threshold of the goal. It encourages the robot to reach the close vicinity of the target. \\
        \bottomrule
    \end{tabular}
\end{table}

\begin{table}
    \centering
    \caption{\small Reward terms for \push~task.}
    \label{tab:reward_terms_push}
    \begin{tabular}{lp{10cm}}
        \toprule
        Reward Term & Expression \\
       \midrule
        \( r_{\text{joint-limits}} \) & \( r_{\text{joint-limits}} = -\max(0, q - q_{\text{upper}}, q_{\text{lower}} - q) \) \\[5pt]
        & \small same as \reach \\
               \midrule
        \( r_{\text{obj-dist}} \) & \( r_{\text{obj-dist}} = -\text{dist}(\text{object}, \text{goal}) \) \\[5pt]
         & \small This term provides a continuous reward that is inversely proportional to the Euclidean distance between the manipulated object and the target goal position. Minimizing this distance maximizes the reward. \\
        \midrule
        \( r_{\text{termination}} \) & \( r_{\text{termination}} = \begin{cases} 1.0, & \text{if } \text{dist}(\text{object}, \text{goal}) < \text{threshold} \\ 0, & \text{otherwise} \end{cases} \) \\[8pt]
         & \small This term offers a one-time, positive reward of 1.0 if the distance between the manipulated object and the goal position falls below a specified threshold, signaling the successful completion of the task. Otherwise, the reward is 0. \\
        \bottomrule
    \end{tabular}
\end{table}

\subsection{Termination Criterion}\label{sec:termination}
Termination conditions vary depending on the task:

\begin{itemize}
    \item \textbf{Reach task:} We use a fixed time horizon. If the EE reaches the goal early, this ensures it remains at the target position, encouraging stable and precise control rather than briefly touching the target.

    \item \textbf{Push task:} Early termination is enabled, the episode ends as soon as the object crosses the goal-side boundary. This signals successful task completion and avoids unnecessary steps afterward.
\end{itemize}

The different termination criteria represent the fact that success detection is straightforward in the push task, allowing the robot to stop immediately once the goal is reached. In contrast, for the reach task, even if the end-effector reaches the goal accurately, the robot often retains residual velocity, making it difficult to stop instantly. 

\section{Experiment Setup}\label{sec:training}

\paragraph{PPO Agent.} We implement a wrapper around the SKRL implementation \cite{serrano2023skrl} to enable agents to work in our multi-morphology setting. During training, we spawn 1280 parallel environments in Isaac-sim, equally divided across the morphologies being trained. For example, when training multi-embodiment baselines with 10 morphologies, each morphology has 128 parallel environments. For single-embodiment training, all 1280 environments belong to the same morphology. This ensures a consistent number of interactions across all baselines. We perform 128 rollout steps before each update, followed by four learning epochs with 16 mini-batches. The PPO clip ratio is set to 0.2, and the agent optimizes the average discounted return with $\gamma=0.99$. 

\paragraph{Curriculum.} The \push~environment has sparser rewards compared to the \reach~environment. This is because, often, the robot swings freely without receiving strong feedback for solving the task. When it does interact with the block, it may push the block in the opposite direction, hindering learning. To simplify learning across multiple morphologies, we apply a curriculum to the \push~task by gradually increasing the goal's $y$-value threshold.   

\paragraph{Compute}
We train all policies on single GPUs (a mix of 3090s and L40s). For 1M steps, MLP policies take approximately 1 day to train, while Transformer policies take around 4 days.

\paragraph{Evaluation.} 
For the \reach~task, the score is the average task reward over 10k evaluation steps. The task reward, $r_{\text{task}}$, is defined as the sum of the EE-to-goal distance penalty $r_{\text{ee-goal}}$ and the vicinity reward $r_{\text{vicinity}})$. The final reported score for \reach~in a given morphology is relative to a random agent: $R = r_{\text{agent, task}} - r_{\text{rand, task}}$.
For the \push~task, the reported score is the success rate--the proportion of episodes where the agent successfully pushes a block to the correct side, averaged over 10k evaluation steps.

\section{Fine-tuning Results}\label{sec:finetuning}
\begin{figure}[H]
    \centering
    \includegraphics[width=\textwidth]{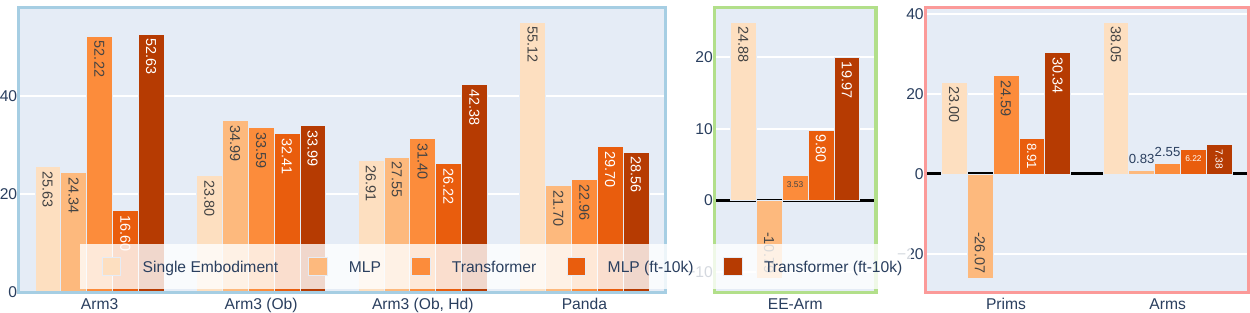}
  \caption{Zero-shot and Finetuned model performance on test morphology for \reach.}
  \label{fig:zs_reach_full}
\end{figure}

\begin{figure}[H]
    \centering
        \includegraphics[width=\textwidth]{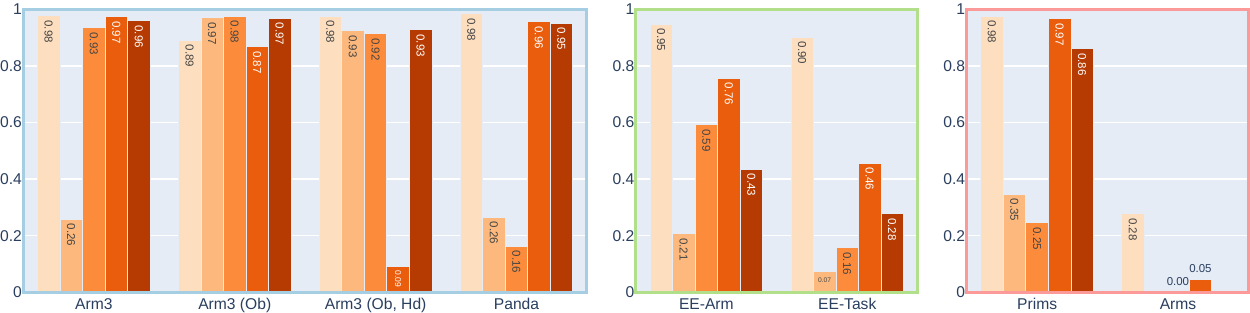}
  \caption{Zero-shot and Finetuned model performance on test morphology for \push.}
  \label{fig:zs_push_full}
\end{figure}

Figure \ref{fig:zs_reach_full} and \ref{fig:zs_push_full}
show the results, including the fine-tuning performance (fine-tuning for 10k steps, each ME agent). We observe that for \textit{Arm3} benchmark tasks, the zero-shot performance of ME agents already surpasses the SE performance, and fine-tuning isn't necessarily needed.  However, in high-dimensional scenarios, fine-tuning does lead to further improvements.

\begin{figure}[H]
    \centering
    \includegraphics[width=0.85\textwidth]{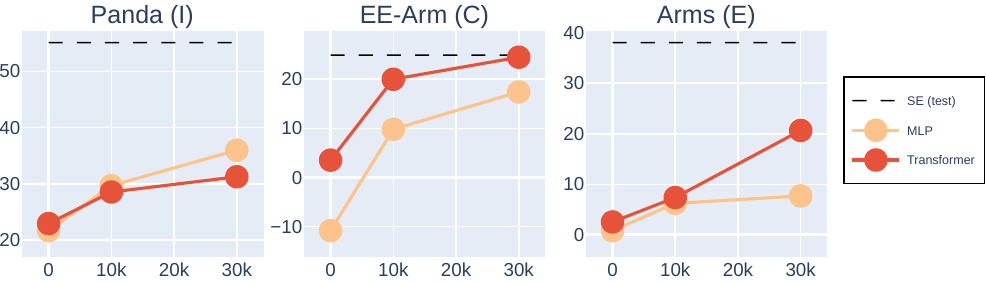}
  \caption{Performance (y-axis) on \reach~task with the number of fine-tuning steps (x-axis).}
  \label{fig:ft_scatter}
\end{figure}

In more complex robot scenarios, fine-tuning is particularly beneficial, though the improvements still remain relatively small in many cases. For \reach~tasks—such as Panda (Interpolation), EE-Arm (Composition), and Arms (Extrapolation)—we also tested fine-tuning for 30,000 steps to better understand how performance changes with the number of fine-tuning steps. The trend is shown in Figure \ref{fig:ft_scatter}. While fine-tuning proves to be much more efficient than training from scratch, suggesting that multi-embodiment training helps the model learn more general control, a significant performance gap remains even after 30k steps of fine-tuning. This indicates that the approach of learning morphology-conditioned policies has not yet fully achieved its goal of being able to control any body.

Note that the performance of agents saturates in the \textit{Arm3} environments. These environments are designed to be simple, both to test models on simpler morphologies and to allow a faster train-test cycle. And while the performance of \textit{Arm3} agents converges quickly, the rate of convergence may still vary significantly, as shown in Figure \ref{fig:convergence_plot} of the main paper. On the other hand, tasks with complex robots highlight the room for improvement in multi-embodiment learning methods.

\section{Future Directions}\label{sec:future}
Future works can focus on the following areas to improve agent learning.

\begin{enumerate}[leftmargin=0.3in, itemsep=0pt, topsep=0pt, partopsep=0pt]
    \item \textbf{Different Policy Architectures.} Our baseline architectures are generic architectures of an MLP and a transformer. However, it is possible to develop specialized architectures that process morphologies in meaningful chunks. Transformers treat each token independently, while MLP concatenates all tokens together. Specialized architectures could break down the morphology into smaller, more relevant sub-units, which might lead to better learning and performance, rather than treating the entire morphology as a single unit.
    \item \textbf{Different Morphology Representation.} 
     We currently approximate link geometries in real robots using basic primitive shape parameters. This approximation might limit the model’s performance. A more accurate representation of morphology could improve the model’s effectiveness.
    \item \textbf{Efficient Morphology-Aware Training.} Our current multi-task objective treats all morphologies equally. However, this approach might not provide the most valuable or optimal information for learning, slowing down the model's learning. 
    Instead, the model should learn to weigh morphologies that are more informative, possibly improving training efficiency and reducing computational costs in reinforcement learning.
    \item \textbf{Cross-Embodiment and Multi-Task} 
    The problem of generalizing to unseen morphologies is addressed using a multi-task objective. It would be interesting to explore whether multi-task learning methods could directly benefit multi-embodiment learning as well.
\end{enumerate}

\end{document}